# EfficientPPS: Part-aware Panoptic Segmentation of Transparent Objects for Robotic Manipulation

Benjamin Alt*, ArtiMinds Robotics, Karlsruhe, Germany
Minh Dang Nguyen*, ArtiMinds Robotics, Karlsruhe, Germany
Andreas Hermann, ArtiMinds Robotics, Karlsruhe, Germany
Darko Katic, ArtiMinds Robotics, Karlsruhe, Germany
Rainer Jäkel, ArtiMinds Robotics, Karlsruhe, Germany
Rüdiger Dillmann, Intelligent Systems and Production Engineering, Forschungszentrum Informatik, Karlsruhe, Germany
Eric Sax, Institut für Technik der Informationsverarbeitung (ITIV), Karlsruhe Institute of Technology, Germany

{benjamin.alt, dang.nguyen, andreas.hermann, darko.katic, rainer.jaekel}@artiminds.com
dillmann@fzi.de
eric.sax@kit.edu
* The authors contributed equally.

## Abstract

The use of autonomous robots for assistance tasks in hospitals has the potential to free up qualified staff and improve patient care. However, the ubiquity of deformable and transparent objects in hospital settings poses significant challenges to vision-based perception systems. We present EfficientPPS, a neural architecture for part-aware panoptic segmentation that provides robots with semantically rich visual information for grasping and manipulation tasks. We also present an unsupervised data collection and labelling method to reduce the need for human involvement in the training process. EfficientPPS is evaluated on a dataset containing real-world hospital objects and demonstrated to be robust and efficient in grasping transparent transfusion bags with a collaborative robot arm.

## 1 Introduction

Aging populations increase demand for hospitals and retirement facilities to provide affordable, high-quality care. At the same time, many societies face growing workforce shortages in the health care sector, making the provision of affordable care increasingly difficult. The use of robots in hospitals enables the automation of many routine tasks, promising to free up qualified staff to focus on more complex, patient-focused tasks that require human skills and judgment. While routine for humans, however, even comparatively simple fetch-and-place tasks in hospital settings can be highly challenging for robots, as they often involve deformable or transparent objects such as syringes, tubes or transfusion bags. Reliable grasping and manipulation of such objects requires robust, intelligent perception systems capable of detecting objects in unstructured environments with variable lighting conditions, while at the same time providing semantic information about the detected objects such as object types or graspable regions.

To that end, we contribute EfficientPPS, a robust and efficient neural architecture for part-aware panoptic segmentation, which simultaneously solves panoptic and part segmentation in one single network. This allows for the detection and annotation of part-whole relationships to provide robots with semantically rich visual information for use in downstream tasks such as grasping and manipulation. In order to address the specific challenges posed by transparent objects in hospital settings, we also contribute a data collection and weakly supervised labelling method to reduce the amount of human involvement in the training process.

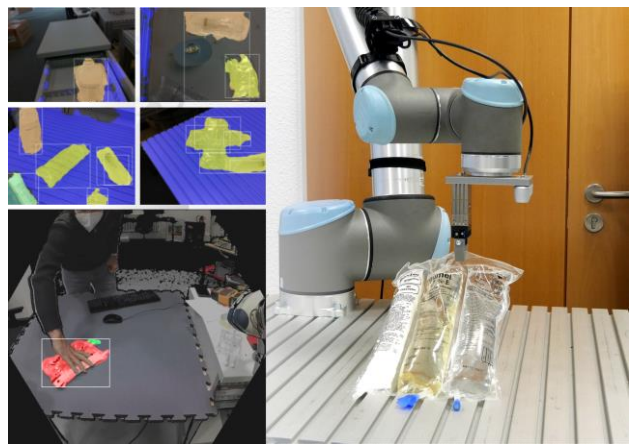

**Figure 1** EfficientPPS enables the part-panoptic segmentation of transparent objects for robotic manipulation in hospital assistance tasks.

Our work is the first to study part and panoptic segmentation of transparent objects. We evaluate EfficientPPS on a dataset containing several real-world hospital objects, as well as in the context of grasping transparent transfusion bags with a collaborative robot arm.

### 1.1 Related Work

#### 1.1.1 Panoptic Segmentation

Originally developed in the context of autonomous driving, panoptic segmentation methods promise robust semantic segmentation of dynamic scenes, while at the same time distinguishing between different instances of the same object class [1]. Early panoptic segmentation approaches performed instance and semantic segmentation separately and combine them in a postprocessing step [1], which leads to considerable computational overhead. More recent models perform instance and semantic segmentation with separate "heads" sharing a common backbone network [2]–[5]. EfficientPS [6] leverages the EfficientNet architecture [7] as a backbone to balance high-precision panoptic segmentation with improved computational efficiency.

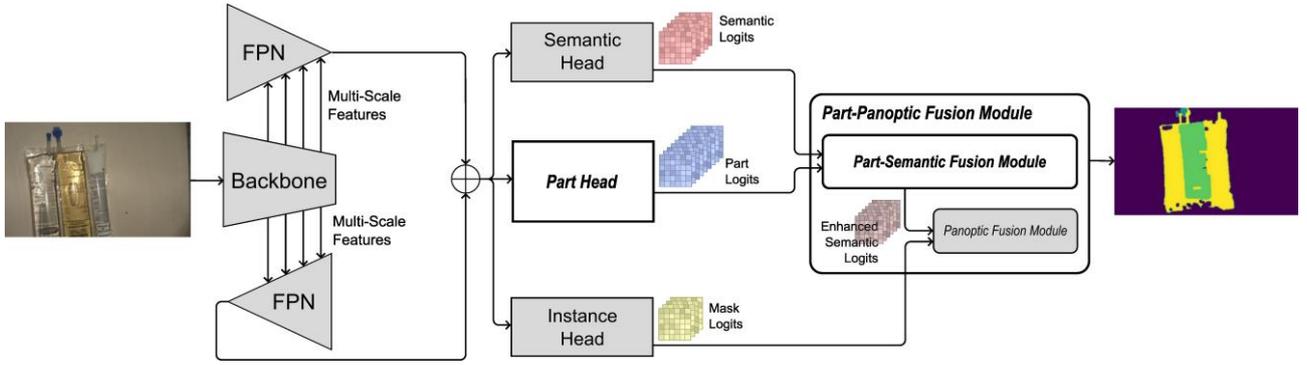

**Figure 2** EfficientPPS architecture. By extending the architecture of EfficientPS by an additional "part segmentation head", panoptic and part segmentation can be performed by a single network. A novel part-panoptic fusion module combines predicted part, semantic and instance information. Our contribution is highlighted in bold.

### 1.1.2 Part-Aware Panoptic Segmentation

Studies in the field of neuropsychology have shown that humans rely on part decomposition to perceive objects [8]. In robotics, part affordances commonly guide the generation of task-dependent grasps [9], particularly in conjunction with part-aware perception methods [10]–[14]. Challenging real-world tasks involving goal-directed manipulation in dynamic environments, such as grasping a transfusion bag in such a way that it can be placed on a hanger, can be solved more robustly with perception methods combining panoptic segmentation with part-aware perception. While the first implementations of part-aware panoptic segmentation [15] used two different networks for panoptic segmentation and part parsing, fusing the results in a postprocessing step, a first unified transformer-based architecture for part-aware perception in the context of autonomous driving has been proposed [16]. We introduce a parameter-efficient unified architecture for part-aware panoptic segmentation based on EfficientPS [6], along with an unsupervised data collection method geared toward robust segmentation of transparent objects.

## 2 Efficient Part-Aware Panoptic Segmentation

We propose EfficientPPS, a novel network architecture for part-aware panoptic segmentation with a single model, as well as an unsupervised training data collection pipeline.

### 2.1 EfficientPPS Architecture

EfficientPPS is a unified neural network architecture for part-aware panoptic segmentation (see *Figure 2*). It builds on EfficientPS [6], a compact and parameter-efficient network architecture for panoptic segmentation. To support simultaneous panoptic and part segmentation, we extended EfficientPS by a *part segmentation head* for identifying part-whole relationships as well as a modified *fusion module* to combine the predicted semantic, instance and part labels.

#### 2.1.1 Shared Backbone, Feature Extraction and Panoptic Segmentation

EfficientPPS uses the same backbone and feature extraction architecture as EfficientPS [6]. A "shared backbone" encoder based on EfficientNet [7] transforms the input image into latent space. The use of EfficientNet's compound scaling permits the backbone encoder to achieve good performance with a comparatively small number of parameters. To effectively learn features at multiple scales, a 2-way Feature Pyramid Network (FPN) [17] is employed. It consists of two branches: A lower branch (see Figure 2) upsamples lower-resolution features to higher resolutions, while an upper branch downsamples higher-resolution features to lower resolutions [6]. To further reduce the number of network parameters, depthwise separable convolutions [18] are used throughout. Instance and semantic segmentation is performed in two separate "heads", which operate on the same internal representation produced by the shared backbone. The instance head is implemented as a variant of Mask R-CNN [19], while the semantic head is based on depthwise separable convolutional layers and Dense Prediction Cells [20].

We refer to [6] for details of the shared backbone, FPN feature extraction as well as the semantic and instance segmentation heads.

#### 2.1.2 Part Segmentation Head

The objective of part segmentation is the classification of pixels of the image into classes. Given an image of several syringes in a drawer, part segmentation associates a class label such as `syringe_barrel`, `syringe_plunger` or `syringe_needle` to each pixel of the image. To perform part segmentation, EfficientPPS adds an additional part segmentation head, which transforms the latent features produced by the feature extraction backbone to pixel-level class labels. For part segmentation, it is not considered which object instance the pixel belongs to. Therefore, part segmentation is a semantic segmentation task, and EfficientPPS uses the same network architecture as its semantic segmentation head for part segmentation.

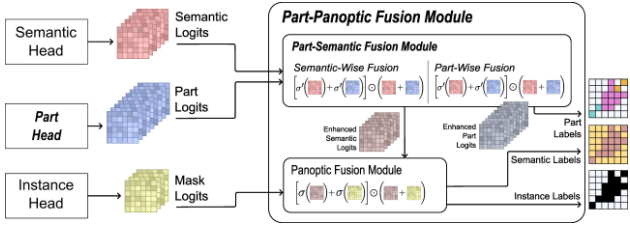

**Figure 3** The part-panoptic fusion module. Part-panoptic fusion is split into distinct part-semantic and panoptic steps, which yield part, instance and semantic labels.

### 2.1.3 Part-Panoptic Fusion Module

Joint part and semantic segmentation poses the fundamental challenge of fusing scene-level and part-level semantics. De Geus et al. [15] show that fusion of part and semantic logits can exploit synergies, as errors made by the part segmentation module can be compensated by correct predictions of a semantic segmentation module and vice versa. However, they also show that requiring both modules to agree on the scene-level label (e.g. syringe_barrel at the part level and syringe at the scene level) decreases the overall predictive performance. To resolve this trade-off, we propose a part-aware panoptic fusion module to fuse scene-level and part-level logits adaptively before integrating instance segmentation information (see Figure 3). The core objective is the amplification or attenuation of logits from different segmentation heads according to an *agreement function*. The fusion module should comply with the following criteria:

**Agreement:** When both heads output logits representing a high probability, the resulting fused logit should be amplified to reflect this consensus.
**Disagreement:** When one head outputs a high probability and the other head a low probability, the fused logit should have a value close to zero.
**Uncertainty:** When one head outputs an uncertain probability, with a corresponding logit close to zero, the fused logit should reflect the logit value of the other head.

The panoptic fusion module of EfficientPS [6] proposes the agreement function

$$L_{SemInst} = \big(\sigma(L_{Sem}) + \sigma(L_{Inst})\big) \odot (L_{Sem} + L_{Inst}),$$

where the fused logit $L_{SemInst}$ is the Hadamard product of the output logits $L_{Sem}$ and $L_{Inst}$ of the semantic and instance heads, respectively. This results in final logits that increase in proportion to the agreement of the semantic and instance heads. We propose to fuse part and semantic information via the following function

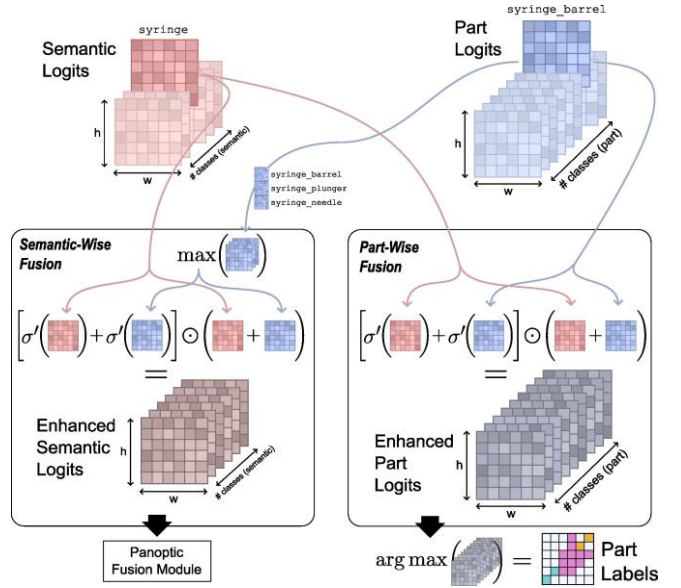

**Figure 4** The part-semantic fusion module. The logits output by the semantic and part heads, respectively, are fused by a semantic-wise and a part-wise routine. Semantic-wise fusion takes a semantic class (e.g. syringe) and all associated part classes (syringe_barrel etc.) and evaluates the agreement function to produce "enhanced" logits for each semantic class, which are forwarded to the panoptic fusion module. Part-wise fusion takes a part class (e.g. syringe_barrel) and its associated semantic class (syringe) to produce "enhanced" logits for each part class, from which the final part labels are computed.

$$L_{PartSem} = \big(\sigma'(L_{Part}) + \sigma'(L_{Sem})\big) \\ \odot (L_{Part} + L_{Sem})$$

where $\sigma'$ is a sigmoid function rescaled to the range [-1, 1]:

$$\sigma'(x) = 2\sigma(x) - 1$$

Our proposed scheme to fuse part and semantic information is shown in Figure 4. Part-semantic fusion is performed in two steps:

1. *Semantic-wise fusion:* For each semantic class (e.g. syringe), the part logits for all corresponding part classes (e.g. syringe_barrel, syringe_plunger etc.) are selected from the outputs of the part head. By computing the maximum across the part class dimension, the part logits are flattened to reflect the probability of a pixel belonging to *any* of the selected part classes. Evaluation of the agreement function yields semantic logits which are "enhanced" with the predicted part information. The enhanced semantic units are passed to the panoptic fusion model to be fused with the instance information.

2. *Part-wise fusion:* For each part class (e.g. syringe_barrel), the corresponding semantic class (syringe) is selected. Evaluation of the agreement function yields part logits which are "enhanced" with the predicted semantic information. From the resulting logits, the final part labels are created by finding the part class with the highest logit value for each pixel.

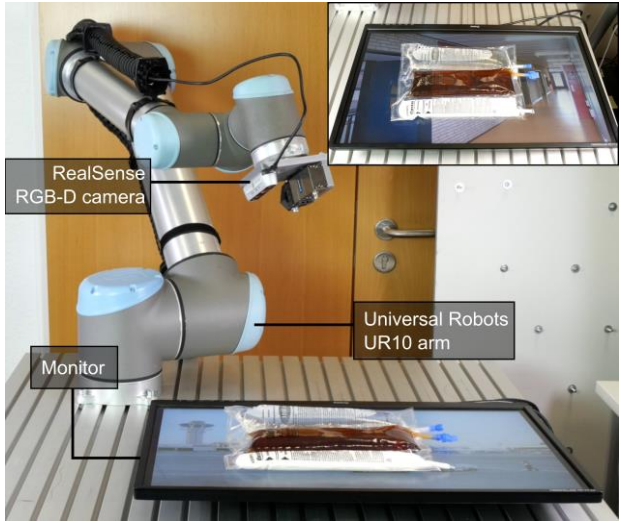

**Figure 5** Hardware setup for unsupervised training data collection. An RGB-D camera is mounted to the flange of a robot arm to collect training data from a variety of perspectives. A monitor is used to generate diverse backgrounds.

The panoptic fusion module is identical to that of EfficientPS [6], with the exception that it uses the "enhanced" semantic logits. The final output of the network are three labels per pixel: The predicted part class, semantic class and instance label.

## 2.2 Unsupervised Data Collection for Transparent Objects

In typical application contexts such as households or hospitals, a large variety of objects must be perceived in various configurations and lighting conditions. In hospitals, transparent or highly reflective objects such as tubes, syringes or transfusion bags are ubiquitous. We propose a data collection pipeline which reduces the need for human labelling while efficiently producing rich datasets for such challenging objects.

We propose to acquire raw image data with a flange-mounted camera and a six-axis robot arm (see Figure 5). By using a robot arm for data acquisition, the capturing angle and position can be varied autonomously to capture the same scene from multiple points of view. We test two variants of the pipeline: Variant A uses an RGB-D camera to use both colour and depth information for training data generation, while variant B uses a 2D RGB images and a monitor to project background textures.

### 2.2.1 Variant A: Label Generation from RGB-D Images

In a first variant of the pipeline, objects are placed directly on a table in the workspace of the robot. An RGB-D depth camera is mounted to the robot flange. To generate labelled training data, we propose a three-step labelling process,

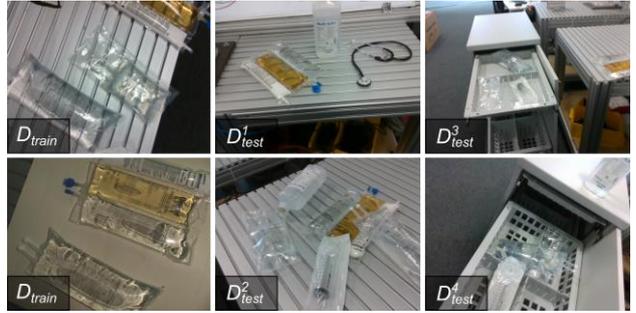

**Figure 6** Examples from the training and test datasets containing transparent objects, with and without overlaps.

which explicitly addresses the challenges posed by transparent objects:

1. *Object labelling:* A progressive morphological filter [21] is applied to the raw point cloud to approximately segment objects from the background. Transparent objects often result in highly noisy depth estimates. For this reason, the segmentation is refined by performing RANSAC [22] with a plane to extract background points missed by the morphological filter. Euclidean cluster extraction is applied to associate the remaining ambiguous points with either the object or the background.
2. *Part labelling:* Because the object classes considered for the experiments (see Section 3) afford colour-based part segmentation, object points are assigned to parts via colour thresholding.
3. *Mapping into pixel space:* The segmented point cloud is projected into the pixel plane using the camera's projection matrix. This projection is not bijective: There may be pixels for which there is no point in the point cloud, or multiple points may correspond to the same pixel. We use a k-NN-based projection algorithm, which treats the projected points as training examples. For each pixel, the nearest $k$ neighbours of the projected points vote for its label.

### 2.2.2 Variant B: Label Generation from 2D RGB Images with Monitor

The appearance of transparent objects changes significantly with the background. We explore a second variant of the data generation pipeline, which employs a monitor to collect datasets with a wide range of backgrounds (see Figure 5) and avoids the need for depth information. For each new camera position, angle, and configuration of objects in the scene, a reference segmentation is performed with two images, one with a blue background and one with a black background. Then, a large number of additional images are taken with randomly selected backgrounds from the DTD dataset [23]. To achieve this, the following steps are performed:

1. Apply morphological closing on each colour channel of the image with the blue background to remove small noises in the colour space.
2. Employ colour quantization to reduce the number of colours in the first image.
3. Perform segmentation in HSV space and locate the outline contour of the object to close all holes inside.

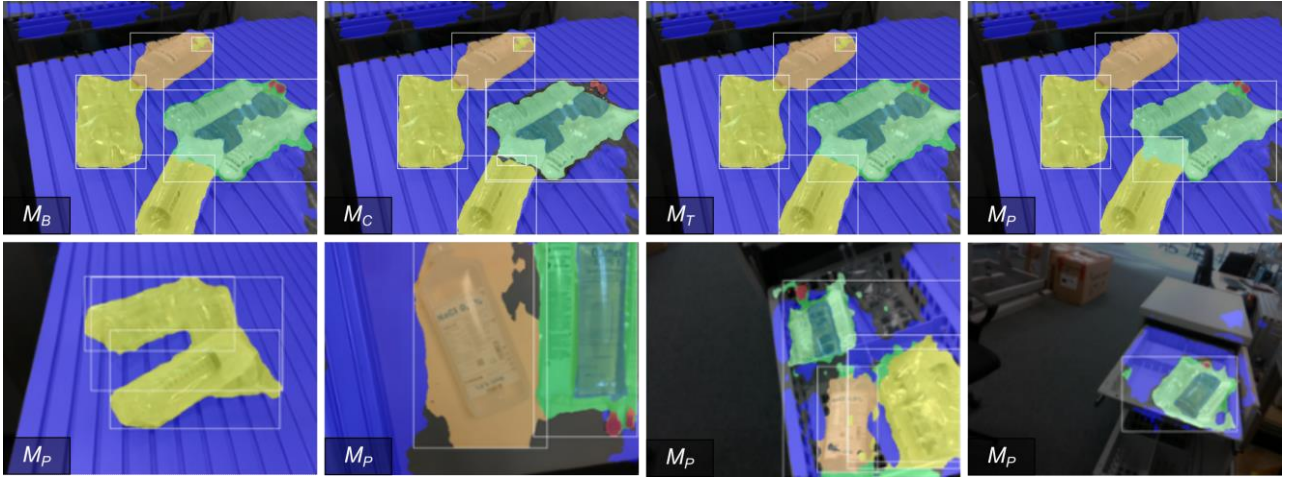

**Figure 7** Top row: Example of segmentation results for EfficientPPS with part-panoptic fusion ($M_P$) and 3 ablations. Part-panoptic fusion ($M_P$) reduces "bleeding" of the `transfusion_bag` (green) into the table (blue) and improves part segmentation, notably for `transfusion_bag_seal` (red). Bottom row: Segmentation results for different scenes featuring overlapping objects, diverse backgrounds and camera perspectives.

4. Extract the mask from the blue-background image.
5. Apply the mask to the black-background image.
6. Perform morphological closing on each channel, apply colour quantization, and perform segmentation in HSV colour space.
7. Locate the outline contour of the object.
8. Repeat the process to identify the part of the object, extracting the reference mask and applying it to other images with random backgrounds.

The resulting training dataset permits the training of models which are highly robust to different backgrounds, a requirement when facing transparent objects. While in principle similar to [24], our pipeline in variant B is the first to leverage a monitor and robot to train a background-invariant (part-)panoptic segmentation model.

## 3 Experiments

To assess the quality of part-aware panoptic segmentation with EfficientPPS, a series of experiments was conducted. We evaluate EfficientPPS on datasets containing partially overlapping transparent, deformable medical objects on various backgrounds and perform ablation studies to determine the benefit of our proposed part-panoptic fusion module. Moreover, we demonstrate the applicability of EfficientPPS to robotic grasping.

### 3.1 Datasets

Using variant A of the data collection method outlined in Section 2.2, we collect one large training dataset $D_{train}$ containing transparent objects from three distinct semantic classes representative of objects encountered in medical assistance settings:

- `transfusion_bag`, a three-chambered transparent bag containing three different types of transfusion fluid in clear, transparent light brown and milky white colours. We define three part classes: `transfusion_bag_seal`, `transfusion_bag_center` and `transfusion_bag_other`, defining the sealing cap, the central chamber and the remainder of the bag, respectively;
- `bottle`, a semi-transparent plastic bottle containing NaCl solution;
- `medical_bag`, a catch-all class for several types of transparent bags of different shapes containing clear transfusion fluid or syringes;
- `table`, the table on which the objects are located.

Due to the requirements of the automatic data labelling pipeline, the training dataset contains only scenes with non-overlapping objects. To increase the diversity of the training data, additional data augmentation was applied (rotational and vertical flipping).

We define four different test datasets containing more complex, realistic scenes. $D_{test}^1$ and $D_{test}^2$ are tabletop scenes without and with overlapping objects, respectively. $D_{test}^3$ and $D_{test}^4$ are objects in cabinet drawers without and with overlapping objects, respectively (see Figure 6). All test datasets contain objects which are partially out of frame and objects not known at training time. Due to the inclusion of overlapping objects in the test data, ground-truth labels for the test datasets are generated via manual annotation.

### 3.2 Quantitative Results

We trained EfficientPPS on $D_{train}$ using EfficientNet-B0 [7] as a shared backbone, initialized with weights pre-trained on ImageNet [25]. EfficientPPS was trained on a single consumer-grade GPU (NVIDIA RTX 2060 SUPER) with mixed precision activated during training [26]. Input images were resized to 480x480 pixels. The network was trained to minimize Part-Panoptic Quality (PartPQ) [15], a part-aware extension of the Panoptic Quality metric [1]. Beyond our proposed model with the part-panoptic fusion module, referred to as $M_P$, we train three ablations reflecting the prior state-of-the-art fusion strategies used in [15]:

- A baseline model $M_B$ where no fusion is performed. This can lead to conflicting labels in the final predictions, such as `bottle` for the semantic channel and `transfusion_bag_seal` for the part channel;
- a model $M_C$ where fusion is performed with the "consensus" strategy, where semantic and part predictions are forced to agree on the scene-level label. Conflicting combinations are set to `void`;
- and a model $M_T$ using "top-down merging", which always keeps the semantic label and sets only the part label to `void` in case of conflicts.

The results are shown in Table 1. The part-aware panoptic fusion module performs equivalently or better than existing strategies in all cases, but improves PartPQ most for multi-part objects and complex scenes ($D_{test}^3$ and $D_{test}^4$). As expected for transparent objects, overlapping scenes and unstructured backgrounds decrease segmentation accuracy. However, in some cases, we observe improved performance even for single-part objects in scenes with overlapping objects. We expect this to be due to the part head having learned some (additional) features which are also beneficial for semantic or instance segmentation.

### 3.3 Qualitative Results

Detection of transparent and deformable objects is a challenging vision problem which lacks comparable benchmarks, making the interpretation of the PartPQ values in Table 1 difficult. Figure 7 visualizes some segmentation results. As shown in the top row, our proposed part-panoptic fusion module not only improves the robustness of part segmentation (see the two red regions for `transfusion_bag_seal`, but also semantic and instance segmentation (see the reduced "bleeding" of the `transfusion_bag` labels into the `table` region). EfficientPPS generally deals well with overlapping objects, as shown in e.g. the bottom left image. It is noteworthy that EfficientPPS can successfully label distinct instances (as visualized with the bounding boxes in Figure 7 even when objects are overlapping, despite having been trained only on scenes without overlaps. As expected, less structured backgrounds such as the drawer are more challenging, particularly in the case of `medical_bag` and `bottle`, and can generate significant bleeding (bottom center-right) or noisy labels (bottom center-left). We found EfficientPPS generally robust against variations in camera angle and distance (see bottom right). Increasing the resolution from 480x480 pixels should further improve results, and particularly avoid spurious detections (bottom right).

To test the robustness of EfficientPPS with respect to different sensor hardware, we tested $M_P$ (trained on images from an Intel RealSense camera) on footage from the Kinect Azure camera of the HoLLiE humanoid assistance robot [27]. Segmentation results were robust against sensor noise, object pose variations and occlusions by humans (see Figure 1 (bottom left) and the companion video).

### 3.4 EfficientPPS for Robotic Grasping

We demonstrate the suitability of EfficientPPS in the context of an assistance robot for hospital assistance tasks. Using EfficientPPS trained on Variant B of the data-collection pipeline described in Section 2.2, we developed a robot skill to grasp transfusion bags in order to hang them. Because the lug for hanging the bag is on the opposite side of the bag than the nozzle (`transfusion_bag_seal`), the grasp point is computed at runtime using the part-panoptic annotation of `transfusion_bag_seal`. We tasked the robot to grasp a transfusion bag in 15 different poses. In each pose, the bag could be grasped successfully after at most 3 trials.

## 4 Conclusion

In this paper, we present EfficientPPS, a parameter-efficient network architecture for part-aware panoptic segmentation, as well as an unsupervised training data collection and annotation method using a flange-mounted camera and a robot arm. We evaluate EfficientPPS on a dataset containing challenging transparent and deformable medical objects and demonstrate its use for robotic grasping.

### 4.1 Discussion and Outlook

EfficientPPS leverages multi-scale and multi-level feature extraction to achieve good instance, semantic and part segmentation in a single network. Its compact size permits it to be trained on a single consumer GPU and run at 5 Hz on consumer hardware, making it ideal for use in low-power embedded systems. To our knowledge, this work is the first to study panoptic segmentation of transparent objects.

| $D_{test}^1$ | table | transfusion_bag | medical_bag | bottle | total |
|---|---|---|---|---|---|
| $M_B$ | 93.5 | 84.1 | 68.8 | 63.4 | 77.4 |
| $M_C$ | 93.5 | 83.2 | 68.8 | 63.4 | 77.2 |
| $M_T$ | 93.5 | 84.3 | 68.8 | 63.4 | 77.5 |
| **$M_P$** | **93.5** | **84.3** | **68.8** | **63.4** | **77.5** |
| $D_{test}^2$ | table | transfusion_bag | medical_bag | bottle | total |
| $M_B$ | 90.6 | 37.3 | 43.9 | 53.4 | 56.3 |
| $M_C$ | 90.6 | 36.3 | 43.9 | 53.4 | 56.1 |
| $M_T$ | 90.6 | 37.9 | 43.9 | 53.4 | 56.5 |
| **$M_P$** | **90.7** | **43.1** | **44.1** | **53.4** | **57.8** |
| $D_{test}^3$ | table | transfusion_bag | medical_bag | bottle | total |
| $M_B$ | - | 66.8 | 18.5 | 35.4 | 40.2 |
| $M_C$ | - | 66.4 | 18.5 | 35.4 | 40.1 |
| $M_T$ | - | 67.1 | 18.5 | 35.4 | 40.3 |
| **$M_P$** | - | **67.4** | **18.5** | **35.4** | **40.3** |
| $D_{test}^4$ | table | transfusion_bag | medical_bag | bottle | total |
| $M_B$ | - | 61.2 | 13.3 | 20.4 | 31.6 |
| $M_C$ | - | 61.0 | 13.3 | 20.4 | 31.6 |
| $M_T$ | - | 61.4 | 13.3 | 20.4 | 31.7 |
| **$M_P$** | - | **61.5** | **13.3** | **20.4** | **31.7** |

**Table 1** Part-aware Panoptic Quality (PartPQ) measured for EfficientPPS with the part-panoptic fusion module (MP) and four ablations on four datasets. Part-panoptic fusion improves PartPQ over alternative fusion strategies particularly (but not exclusively) for objects with multiple parts.

The proposed data collection pipeline enables the collection of training datasets without requiring human labelling. The use of both depth and colour information to generate labels is a promising approach for transparent objects, which are notoriously difficult to detect via either modality in isolation. The use of a monitor to generate autonomously labelled datasets of transparent objects with a wide range of backgrounds is promising, as it enabled the training of EfficientPPS at a sufficient segmentation quality to enable robotic grasping, while avoiding the need for depth information at data collection time.

It its current iteration, EfficientPPS showed limited performance on overlapping objects and more unstructured environments, rendered particularly challenging with transparent objects. Further evaluation with larger datasets, a wider range of objects and inclusion of overlapping objects in the training data is required to determine the precise cause of this limitation. We will conduct a quantitative comparison between variants A and B of the proposed data collection pipeline to assess the respective impacts of 3D information or the use of a monitor for data collection. Moreover, future work will evaluate and compare the performance of EfficientPPS on large-scale standard datasets such as COCO [28] or CityScapes [29] and compare its performance with larger models [16].

## 5    Acknowledgments

This work has been partly funded by the German ministry of education and research (BMBF) as part of the HoLLiECares project (reference no. 16SV8406).